\documentclass{article}

\usepackage[preprint]{nips}

\usepackage[utf8]{inputenc} 
\usepackage[T1]{fontenc}    
\usepackage{amsfonts}       
\usepackage{nicefrac}       
\usepackage{microtype}      

\usepackage{graphicx}
\usepackage{subcaption}

\usepackage{amsmath}
\usepackage{amssymb}

\usepackage{algorithmic}
\usepackage{algorithm}
\usepackage{authblk}

\title{Reannealing of Decaying Exploration \\Based On Heuristic Measure in Deep Q-Network}

\author{\textbf{Xing Wang}, \quad
\textbf{Alexander Vinel}\thanks{Corresponding to: alexander.vinel@auburn.edu} }

\affil{Department of Industrial Engineering\\ Auburn University\\ Auburn, AL, 36849, USA}

\begin{document}
\maketitle

\begin{abstract}
Existing exploration strategies in reinforcement learning (RL) often either ignore the history or feedback of search, or are complicated to implement. There is also a very limited literature showing their effectiveness over diverse domains. We propose an algorithm based on the idea of reannealing, that aims at encouraging exploration only when it is needed, for example, when the algorithm detects that the agent is stuck in a local optimum. The approach is simple to implement. We perform an illustrative case study showing that it has potential to both  accelerate training and obtain a better policy.
\end{abstract}







\section{Introduction}
The goal of a reinforcement learning agent is to try to make best decisions, based on the information it gathers along the way. Unlike supervised learning tasks, however, the agent can only have access to the environment through its own actions. It needs to explicitly explore its environment and gather information for decision making. Simultaneous exploitation (making best decisions) and exploration (gathering of information) tasks create a dilemma, and balancing the two is one of the core challenges in reinforcement learning. An early survey \cite{thrun1992} of exploration strategies made a  distinction between two categories: undirected and directed. The key idea behind the former is to add randomness, in the hope that a random action might lead towards better actions compared to the suboptimal policy which is viewed as the best given current information. 
Directed strategies take a ``global'' view and measure some  statistics of the past experiences, and utilize these measures to guide efficient exploration mainly by adding an exploration bonus to the reward function, 
so that the less visited (in terms of pseudo-count \cite{bellemare2016unifying} using a fitted density model or hash-count \cite{tang2017exploration} with locality sensitive hashing) state-action pairs, or those with larger information gain (\cite{chen2017infogain}, VIME \cite{houthooft2016vime}) or prediction error \cite{stadie2015incentivizing},
are favored. 
Such strategies often allow for theoretical analysis, usually based on multi-armed bandit (MAB) problem theory \cite{burtini2015survey}.
In spite of their appealing mathematical formalism and theoretical guarantees in finite case, directed exploration strategies have not shown effectiveness over domains and thus have not played an important role in the recent success of reinforcement learning \cite{tang2017exploration}. 

In many real-life learning applications, RL agents are trained to achieve optimal performance in certain specific tasks. Hence, it is often simple to distinguish when the learned behavior of the agent is acceptable. 
It is well described in the literature that if insufficient focus has been placed on exploration, then the agent can learn to  stay in a ``comfort zone''  of a local optimum. In this case, one needs to force the agent to leave the ``comfort zone'' and try new actions which would take it to new states that have not been well-learned yet, so that it can explore more information about the environment, in the hope of finding better policy. 
In this paper, we propose a  heuristic to overcome this problem, and in general, to speed up learning procedure. 

Our main contribution is an easy yet efficient and scalable method that encourages exploration in complex reinforcement learning domains when it is needed. To be more specific, we emphasize the model dynamics and the agent's behavior rather than uncertainty estimates while measuring the need for exploration. This can be accomplished by training a supervised model to make predictions based on existing experiences, which would require extensive computation as well as the effective representation of the supervised model. In this paper, we focus on using a simple heuristic measure and an annealing-based method to redirect the agent. Our approach could be extended to serve as a general framework for interactively training in complex RL domains, to aid the agent in finding better policies. 

Another contribution is that we abstract the learning procedure with the view of general optimization/search, and apply a generic method that attempts to improve search algorithms on hard problems, specifically a modified version of  simulated annealing and thus our method is referred to as \textit{exploration reannealing}. A number of other metaheuristic approaches can be applied in similar fashion to better balance the exploration-exploitation tradeoff in reinforcement learning.

It is worth noting that exploration reannealing itself is not a complete learning algorithm, and in fact needs to be combined with other reinforcement learning tools. In this paper, we emphasize and evaluate its application to deep Q-learning, but the combination of exploration reannealing with other methods can be  considered.

The paper is organized as follows. Section \ref{sec:bg} provides some background on reinforcement learning in the context of Markov Decision Process (MDP) models, in particular the Q-learning algorithm, deep Q-networks (DQN) and some recent techniques we  exploit to solve DQN. 
In Section \ref{sec:reaneal}, we present the motivation of our reannealing method and discuss the appropriate ways to use it, as well as the specific algorithm. 
We perform empirical studies and showcase the improvement by exploiting our reannealing strategy on a large scale challenging domain, the Lunar Lander model, in Section \ref{sec:res}. 
Finally, we outline conclusions in Section \ref{sec:conclude}.

\section{Background and Preliminaries} \label{sec:bg}

\subsection{Q-learning}
A natural abstraction for many sequential decision-making problems is to model the system as a \textit{Markov Decision Process} (MDP) \cite{puterman1994mdp}, in which the agent interacts with the environment over a sequence of discrete time steps. 
It is often represented as a 5-tuple:  $M=<\mathcal{S}, \mathcal{A}, T, R, \gamma>$, where $\mathcal{S}$ is a set of \textit{states}; $\mathcal{A}$ is a set of \textit{actions} that can be taken; $T: \mathcal{S}\times \mathcal{A} \mapsto \mathcal{P_S}$ is the \textit{transition function} such that $\int_{s'\in\mathcal{S}} T(s'|s, a) = 1$, which denotes the (stationary) probability distribution over $\mathcal{S}$ of reaching a new state $s'$, after taking action $a$ in state $s$; $R$ is the reward function, which can take the form of either 
$R:\mathcal{S} \mapsto \mathbb{R}$, $R:\mathcal{S}\times \mathcal{A} \mapsto \mathbb{R}$, or $R: \mathcal{S}\times \mathcal{A}\times \mathcal{S} \mapsto \mathbb{R}$; and $\gamma \in [0, 1)$ is the \textit{discount factor}.

A policy $\pi: \mathcal{S} \mapsto \mathcal{P_A}$ defines the conditional probability distribution of choosing each action while in state $s$. For an MDP, once a stationary policy is fixed, the distribution of the reward sequence is then determined. Thus to evaluate a policy $\pi$, it is natural to define the \textit{action value function under $\pi$} as the expected cumulative discounted reward by taking action $a$ starting from state $s$ and following $\pi$ thereafter:
\begin{align}
  Q^\pi (s, a) \equiv \mathbb{E}_\pi \Big[\sum_{\tau=0}^\infty \gamma^\tau R_{t+\tau} |S_t = s, A_t = a \Big]  
     = R(s, a) + \gamma \int_{s'} T(s'|s, a) Q^\pi (s', \pi(s')). 
\end{align}

The goal of solving an MDP is to find an \textit{optimal policy} $\pi^\ast$ that maximizes the expected cumulative discounted reward in all states. The corresponding \textit{optimal} action values satisfy $Q^{\ast} (s, a) = \max_\pi Q^\pi (s, a)$, and Banach's fixed-point theorem ensures the existence and uniqueness of the fixed-point solution of \textit{Bellman optimality equations} \textbf{\cite{puterman1994mdp}}: 
\begin{equation}
Q^\ast (s, a) = R(s, a) + \gamma \int_{s'} T(s'|s, a) \max_{a'} Q^\ast (s', a') 
\label{eqn:bellman}
\end{equation}
from which we can derive a deterministic optimal policy by being greedy with respect to $Q^\ast$, i.e., $\pi^\ast = \textrm{argmax}_{a\in \mathcal{A}} Q^\ast (s, a)$. 

In reinforcement learning problems, the agent must interact with the environment to \textit{learn} the information about the transition and reward functions, meanwhile trying to produce an optimal policy. While interacting with the environment, at each time step $t$, the agent senses some representation of current state $s$, selects an action $a$, then receives an immediate reward $r$ from the environment and finds itself in a new state $s'$. The \textit{experience tuple} $<s, a, r, s'>$ summarizes the observed transition for a single step. Based on the experiences through interacting with the environment, the agent can either learn the MDP model first by approximating the transition probabilities and reward functions, and then plan in the MDP to obtain an optimal policy (this is called the \textit{model-based} approach in reinforcement learning); or without learning the model, directly learn the optimal value functions and upon which the optimal policy is derived (this is called the \textit{model-free} approach). 

We use Q-learning with function approximation in this paper. 
As a \textit{model-free} approach, Q-learning \cite{watkins1992q} updates one-step bootstrapped estimation of Q-values from the experience samples over time steps.  
The update rule upon observing $<s, a, r, s'>$ is
\begin{equation}
    Q(s, a) \leftarrow Q(s, a) + \alpha \big( r + \gamma \max_{a'} Q(s', a') - Q(s, a) \big)
\end{equation}
in which $\alpha$ is the learning rate,
$r+ \max_{a'} Q(s', a')$ serves as the update target of the Q-value, which can be seen as a sample of the expected value of one-step look-ahead estimation for state-action pair $(s, a)$, based on the the maximum estimated value over next state $s'$, 
and the last term $Q(s, a)$ is simply the current estimation. 
The difference $\delta = r + \gamma \max_{a'} Q(s', a') - Q(s, a)$ is referred to as temporal difference (TD) error, or Bellman error. 
Note that one can bootstrap more than one step when estimating the target, often by using the \textit{eligibility trace} as in $TD(\lambda)$ \cite{sutton1988td}.
Q-learning is guaranteed to converge to the optimal values in probability as long as 
each action is executed in each state infinitely often, $s'$ is sampled following the distribution $T(s, a, s')$, $r$ is sampled with mean $R(s, a)$, variance is bounded and given appropriately decaying $\alpha$. 

For environments with large state spaces, the Q-values are often represented by a function of state-action pairs rather than the tabular form, i.e., $Q_\theta(s, a) = f(s, a|\theta)$, where $\theta$ is a parameter vector. To update parameter vector $\theta$, first-order gradient methods are usually applied to minimize the mean squared error (MSE) loss:
$
    \theta \leftarrow \theta + \alpha \delta \nabla_{\theta} Q_\theta.
$
However, with function approximation, the convergence guarantee can no longer be established in general. Neural networks, while attractive as a powerful function approximator, were well known to be unstable and even to diverge when applied for reinforcement learning until deep Q-network (DQN) \cite{dqn} was introduced to show great success, in which several important modifications were made. 
\textit{Experience replay} \cite{lin1992experiencereplay} was used to address the non-stationary data problem, by storing and mixing the samples (i.e., experiences) into a replay memory for the updates. During training a batch of experiences is randomly sampled each time and the gradient descent is performed on the sampled batch. This way the temporal correlations could be alleviated. In addition, a separate \textit{target network}, which is  a copy of the learned network parameters ($\theta$) is employed. This copy is frozen for a period of time and is only updated periodically (denoted as $\theta^-$), and is applied to calculate the TD error, with the aim of improving stability. 
A variety of extensions and generalizations have been proposed and shown successes in the literature.
Overestimation due to the max operator in Q-learning may significantly hurt the performance. To reduce the overestimation error, double DQN (DDQN) \cite{ddqn} decouples the action selection from estimation of the target, that is, choosing the maximizing action according to the original network ($Q_\theta$), and evaluate the current value using the other one ($Q_{\theta^-}$ from the target network), i.e., 
$
Q_\theta (s, a) \leftarrow r + \gamma Q_{\theta^-} (s', \textrm{arg}\max_a Q_\theta (s', a)).
$
Interested readers are referred to \cite{wang2020advances} for further reading of more DQN architectures.
In this paper, unless stated explicitly, we use the DDQN update for all DQN architectures. 


\subsection{Exploration Strategies}
\paragraph{$\varepsilon$-Greedy Exploration.}
The most commonly-used strategy for exploration is the $\varepsilon$-greedy method, in which the agent selects the action it believes to be the best according to current $Q(s, a)$ values for the most of the time, and occasionally acts randomly. That is, it takes the greedy action $\textrm{arg}\max_\pi \int Q(s, a)\pi(a|s)da$ with probability $1-\varepsilon$, and selects (uniformly) randomly among all actions with probability $\varepsilon$. Then after infinitely many steps, every state-action pair will be visited infinitely often, thus all $Q(s,a)$ converge to the true action values $Q^\ast (s, a)$ almost surely \cite{suttonbook}. However, deficiencies of $\varepsilon$-greedy are also often discussed and new RL algorithms can be proposed. For example, the time complexity of $\varepsilon$-greedy learning is exponential with respect to the size of the state space, which leads to PAC-learning ideas for RL \cite{strehl2009pac}. Moreover, $\varepsilon$-greedy selects actions with equal probability. Intuitively, we would expect the agent to pay more attention to more ``promising'' actions, i.e., those with maybe slightly lower $Q$-values than the current greedy action, rather than those with really low $Q$-values, which have less potential to be optimal. Moreover, it might be a waste to explore those actions with low $Q$-values, which have been selected many times since we may be confident that these are ``bad'' actions.
Despite its deficiencies, due to its simplicity, practical effectiveness, and the ease with which it can be  embedded into Q-learning, $\varepsilon$-greedy strategy has been prevalent in most value-based algorithms in reinforcement learning, including DQN and its variants.

\paragraph{Softmax (or Boltzmann) Exploration.}

The (variational) free energy for an RL agent can be defined as 
\begin{equation}
F(\pi) = - \int Q(s, a)\pi(a|s) da + T \int \pi(a|s) \log \pi(a|s) \textrm{d}a,
\label{eqn:energy}
\end{equation}
in which the first term represents the energy of the agent, and the second term is the standard form of negative entropy. Coefficient $T$ of the negative entropy is referred to as \emph{temperature}. 
Free energy principle claims that a self-organizing agent would act on the environment by minimizing its free energy, by which it reaches an equilibrium with the environment (or more accurately, a sampling of sensory data) \cite{free_energy_nature}.
Minimization of free energy gives us 
\begin{equation}
\pi(a|s) = \dfrac{ \exp \big( \frac{Q(s, a)}{T}\big) }{\int \exp \big( \frac{Q(s, a')}{T} \big) \textrm{d}a'}
\label{eqn:softmax}
\end{equation}
which is called the softmax policy or Boltzmann policy. 
Note that hyperparameter $T$ controls the exploration \cite{ishii2002control}. If the temperature is high, the action selection according to $\pi$ approaches uniform distribution, which yields more randomness and thus encourages exploration. On the other hand, low temperature would reduce the randomness and enhance exploitation. As an extreme case, if the temperature is zero, the negative entropy term in Eq. (\ref{eqn:energy}) goes away, and the corresponding policy becomes deterministic which takes the greedy action 
$\textrm{arg}\max_\pi \int Q(s, a)\pi(a|s)da$ given the current estimate of $Q(s, a)$.
With the softmax probability, the possibilities for each action to be selected are ranked and weighted relevant to their estimated $Q$-values, instead of equal probabilities for all actions in $\varepsilon$-greedy approach. 

\subsection{Exploration Decay}
Theoretical analysis of exploration strategies is usually performed through the Multi-Armed Bandit (MAB) model \cite{burtini2015survey}. $\varepsilon$-Greedy strategy has been well studied through regret analysis in MAB, in which the regret is often defined as a measure of the difference in value between taking an action $a$ and the optimal action $a^\ast$ at time $t$, i.e., $\rho(t) = \mathbb{E}_a[V^\ast - Q(a_t)]$, that is, the opportunity loss of taking $a_t$ for one step. The total regret is then the overall opportunity loss over time until time $t$, i.e., $ L(t) = \sum_{\tau=1}^t \rho(\tau) = \mathbb{E}_a[\sum_{\tau=1}^t (V^\ast - Q(a_\tau))].$ As shown in \cite{burtini2015survey}, if we set $\varepsilon=0$, that is, always choose the action with the largest $Q$-value greedily without exploration attempt, then the greedy action could lock onto a suboptimal policy forever, in that case, a linear bound ($\mathcal{O}(t)$) on total regret is achieved. On the other hand, if we take $\varepsilon$-greedy action with constant $\varepsilon > 0$, the agent would keep exploring with probability $\varepsilon$ even if the optimal policy is found, thus also resulting in linear total regret. 

A natural approach, then, is to encourage  exploration early and exploitation later, which is achieved with decaying $\varepsilon$ over time. 
Decaying-$\varepsilon$-greedy can achieve asymptotically logarithmic bound on total regret, by defining $\varepsilon_t = \min(1, \frac{c}{\delta^2 t})$, where $c$ is a constant and $\delta$ is the gap between the best and second best action values, both are unknown however. Thus it is often hard to derive an efficient decay schedule. Nevertheless, it is important to emphasize the decay strategy on exploration. We note that the exploration of stochasticity for softmax strategy could also be annealed during training by changing the temperature $T$ over time. 

\section{Exploration Reannealing} \label{sec:reaneal}

\subsection{Local Optima in DQN}

In theory, $Q$-learning converges to the optimal policy if all state-action pairs are visited infinitely often. However, this condition cannot be met in practice if the state-space  is too large or continuous.
A neural network in DQN approximates large or continuous state space and thus suffers from this problem. 
A deep neural network in general is of very high dimensinality, and popular practical optimization techniques, such as stochastic gradient descent, only consider first-order gradient information of the loss function. Such optimization algorithms may get stuck at local optima or saddle points.
In practice, for regular neural networks, saddle points of the loss function  can be escaped by applying special  optimization techniques \cite{anandkumar2016saddle}, and it is often the case that a local optima is good enough for many supervised learning problems. However, this might not be the case in DQN. A local optimum in DQN arises from both  the complicated structure of loss function itself, and   the limited representation and inference ability of a neural network for the search space. The latter is especially pervasive for state-space segments that are not well-explored.  As a result, the learning agent cannot make progress for  a long time and might waste learning resources by  updating information for  irrelevant parts of state space. 
Therefore, it is more common in practice that a DQN learning agent gets stuck in poor local optima, due to the difficulty of handling the exploration-exploitation trade-off well.

\subsection{Exploration Reannealing}
Simulated annealing (SA) is a classic heuristic optimization approach used to escape local optima. At the heart of it is an analogy with thermodynamics. Boltzmann probability distribution again is used to analogically represent the (variational) free energy, with a control hyperparameter, referred to as  temperature $T$. The free energy determines the stochasticity for the search direction, which aids the local search to escape from local optima. The concepts are fundamentally based on the same principle as those in exploration strategies we mentioned above, in which decaying the exploration is fundamentally the same as tuning the temperature in SA. 
In some variations of simulated annealing search, re-anneaing \cite{ingber1989very} is a quite common idea for the anneal schedule, that is, the temperature is periodically set to a high value in order to encourage exploration. 

Similar idea can be naturally employed in our problem for enabling exploration in RL. Note that the act of  reannealing itself can be implemented in a straightforward way for both $\varepsilon$-greedy and softmax strategies we introduced above. In $\varepsilon$-greedy, we can easily reset the exploration rate $\varepsilon$ to a large value (note that $0\le \varepsilon < 1$ and $\varepsilon$ decays over time) if poor local optima is met. Similarly, for softmax action selection, we can more directly reset the temperature to a high value (e.g., close to the initial temperature) whenever it is necessary. 
We note here that with finitely many reannealing events, the theoretic guarantee of asymptotically logarithmic bound on total regret will still hold as for decaying $\varepsilon$-greedy.  A more significant challenge  relates to the timing  of reannealing events. We will discuss this question below, but first we discuss additional reasons why we believe reannealing can bring benefits for learning in DQN.

The key advantage of reannealing exploration is that it could substantially improve the sample efficiency. 
We know that collecting data by interacting with the environment is usually expensive for RL systems.
While stuck in local optima, 
it is usually the case that the TD errors being backpropagated are small, and the agent could learn little information thus gain little learning progress. With reannealing, the agent would tend to take random actions in this case and is more likely to experience  unacquainted states thereafter. Those state-action pairs are usually visited much less often than those obtained by taking greedy policy, thus have larger TD errors in general and from which the agent can learn more.

Another advantage is that reannealing exploration could substantially alleviate the data imbalance problem.
Without reannealing exploration, large amounts of samples are collected around local optima, resulting in data distribution biased in favor of samples that may not be relevant. As a result, a notable portion of  model parameters are dedicated to describing states around (poor) local optima, and much of the training work is hence in vain. 
By reannealing the exploration instead of exploiting around the local optima, random actions are taken with much higher probabilities, the agent are more likely to jump out of the local optima and experience with unacquainted states, gather significantly more useful information about the entire environment as well as the training overall.

Finally, a training episode is often designed to have a finite horizon for computational simulation purpose. Each episode finishes when either certain criteria are met (in this case, a success or a failure on the task is defined and final reward is given), or the time step exceeds a fixed period. When a local optimum is encountered, the agent tends to wander around until exceeding the time limit of an episode. It is important to note that using a time limit makes the environment non-stationary, since in this case the final reward is never actually assigned, and hence, the agent may not be able to recognize a suboptimal policy.  Exploration reannealing can enable  the agent to actually achieve either success or failure, making sure that appropriate reward is assigned. Consequently, more episodes finish with more concrete information gain.


\subsection{Defining Poor Local Optima}

Given the  intuitive advantages of applying reannealing to DQN, we next describe our proposed algorithm. As described above, the mechanism of reannealing is straghtford for both $\varepsilon$-greedy and softmax strategies. On the other hand, detemining the appropriate times for reannealing is more difficult.
Clearly, we must reanneal, when the agent is stuck in a poor local optimum. 
Unfortunately, in high-dimensional spaces formally determining local optimum is challenging. In practice, an often used empirical way is to track variation of loss function  across iterations. When the loss stops improving, it is often the case that the search reaches local optimum. However, simply looking at the change in loss does not tell us whether the local optimum is acceptable. It might be the case that near-global optimum has already been achieved, and hence there is no need to escape from it.

On the other hand, sometimes a poor local optimum can be easily observed and distinguished by human from the outside perspective, in which case the observer utilizes some \textit{a priori} knowledge that has not been integrated into the reward function. In RL, the agent's learned policy as well as its behavior are determined by optimizing the discounted cumulative rewards, thus an ideal reward function should capture the goal and measure the performance exactly, which requires perfect knowledge of all states and transitions in the environment. 
Except for some human designed games in which the rules are entirely understood, it often takes considerable effort to tweak the rewards until desired behavior is learned. This then means, that in many applications the reward function is already overloaded in a way as to result in favorable agent's behavior, and, hence, attempting to also use the same reward function to distinguish the quality of local optimum may be either impossible or very complicated with unexpected side-effects (see also inverse reinforcment learning, \cite{ng2000irl}).

Alternatively, we propose to consider a separate criterion for initiating reannealing. This criterion can be viewed as a supervised model, which makes predictions based on existing experiences. The training labels could be as simple as a categorical signal to denote the need to explore, or as complicated as representation of next state, which would require extensive computation as well as the effective representation of the supervised model.
In this paper, we use a simplified version of this supervision idea. We explicitly measure the easily distinguished feature as an a priory defined \emph{heuristic}, representing the fact that the agent's bottleneck behavior due to sub-optimal policy can often be described with  some undesirable characteristics from an outside observer's perspective. 
We, then, can explicitly extract such a feature as a useful heuristic independent from the reward function.
Once defined, we can keep track of the heuristic along the learning process and use it to control the learning behavior. See Section \ref{sec:res} for an example based on Lunar Lander problem.

\subsection{Algorithm}
The objective of reannealing exploration is to explicitly inform the learning agent that it should be exploring rather than exploiting with a heuristic measure. 
We set up a heuristic variable called \texttt{stuck} to represent if the agent has been stuck in poor local optima. The variable \texttt{stuck} should be a global statistic for some aspect of the agent's performance information.
If some threshold of ``stuck'' has been reached, we reanneal the exploration. In $\varepsilon$-greedy learning, we reset $\varepsilon$ to 1 and force the agent do pure exploration. The exploration rate $\varepsilon$ then is decayed over time. The pseudo-code of our proposed procedure for DQN is shown in Algorithm \ref{alg:reanneal}. And similar reannealing strategy applies for softmax, in which we reset the ``temperature'' $T$ to its initial value, and anneal it again to smaller value over time. 

\begin{algorithm}[htb!]
\caption{Exploration Reannealing in DQN}
\label{alg:reanneal}
\begin{algorithmic}
\STATE Initialize $\varepsilon_t = 1$ and \texttt{stuck}, as well as DQN parameters $\theta$.
\REPEAT[for each episode]
    \STATE initialize state $s$
    \REPEAT [for each step in an episode]
        \STATE Generate a random number $u \in [0, 1]$
        \IF{$u < \varepsilon_t$}
            \STATE Randomly select $a\in A$
        \ELSE
            \STATE $a \leftarrow \textrm{arg}\max_{a'\in A} Q(s, a' | \theta)$
        \ENDIF
        \STATE Take action $a$, observe reward $r$ and next state $s'$
        \STATE Store experience tuple $<s, a, r, s'>$ into memory
        \STATE Sample a batch of experiences from memory
        \FORALL{sampled experience in the batch} 
            \STATE Compute the TD error $\delta = r + \max_{a'} Q(s', a'|\theta) -Q(s, a|\theta)$
            \STATE Backpropagate $\delta$ through the DQN, update $\theta$ with learning rate $\alpha_t$
        \ENDFOR
        \STATE $s\leftarrow s'$
    \UNTIL{$s$ is terminal state}
    \STATE Update \texttt{stuck} according to performance
    \IF{\texttt{stuck} meets some threshold}
        \STATE Reset $\varepsilon_t = 1$
        \STATE Reset \texttt{stuck} to its initial value
    \ELSE
        \STATE Decay $\varepsilon_t$
    \ENDIF
\UNTIL{end of learning}
\end{algorithmic}
\end{algorithm}

As argued above, the candidates of the \texttt{stuck} variable should be some performance measure that might have not been integrated (or not been integrated well) in the reward function. The chosen feature as the explicit heuristic should be a representative bottleneck for learning.
We expect that the RL agent could jump out of the local optima by applying reanneal strategy, and be able to learn better policy than the one it obtained before reannealing when it stucks. As a result, acceptable behavior and good policy could be learned faster. We also expect that with reannealing exploration, we could worry less about poor local optima and spend less time on tuning the hyperparameters (such as the annealing schedule, learning rate, etc.) while training. 


\section{Experimental Results} \label{sec:res}

\subsection{Testbed Setup}
\begin{figure}[htb]
    \centering
    \includegraphics[width=.5\textwidth]{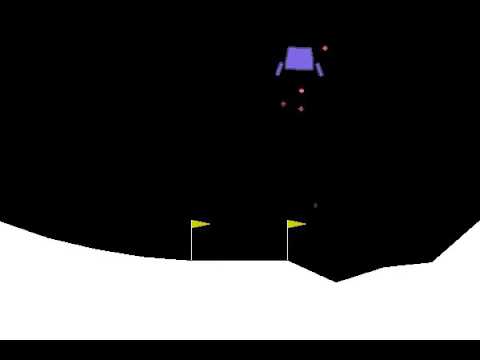}
    \caption{Lunar Lander Environment}
    \label{fig:lander}
\end{figure}

We conducted an experiment by implementing a reinforcement learning agent to solve the Lunar Lander task in Box2D \cite{catto2011box2d}, interfaced through OpenAI gym environment \cite{gym}. In each step, the agent is provided with the current state $s$ of the lander in $\mathbb{R}^8$, in which 6 of the dimensions are in continuous space whereas the other 2 are dummy variables in discrete space, and the agent is allowed to make one of the 4 possible actions (i.e., the action space is discrete). At the end of each step, the agent receives a reward and moves to a new state $s'$. An episode finishes if the lander rest on the ground at zero speed (receives additional reward of $+100$), or hits the ground and crashes (receives additional $-100$ reward), or flies outside the screen, or reaches the maximum of 1000 time steps of one episode. The agent aims for successful landing which is defined as reaching the landing pad (between two flags) centered at the ground at the speed of zero, and receives an additional reward in range  $[100, 140]$, while landing outside the pad would cause some penalty. Figure \ref{fig:lander} provides a snapshot of the task environment.

We use a neural network with two fully-connected hidden layers (which consist of 200 and 60 neurons, respectively) as our function approximator. ReLU nonlinearity is utilized as the activate function for each hidden neuron. The network takes the 8-dimensional vector $s$ which describes the state as the input, and outputs the approximated $Q$-values for the 4 possible actions. 
We train the neural network with a FIFO memory of size $10^6$ for experience replay. A target neural network for double learning is updated every 20 episodes, so that the original network has enough time to converge.
The adaptive moment estimation (Adam) optimizer with initial learning rate set to 0.01 is used to train the network, since it is in general less sensitive to the choice of the learning rate than other stochastic gradient descent algorithms \cite{kingma2014adam}.
We apply the pseudo-Huber loss instead of MSE as the loss function, as it is less sensitive to outliers and is more commonly used in DQN \cite{dqn}. The discount factor $\gamma$ is set to 0.99, and $\varepsilon$-greedy policy is used for choosing actions throughout interacting with the environment. For comparison purpose, we used two different exploration decay rates, $\rho_{decay}=0.99$ and 0.985. These hyperparameters are empirically tuned in the aim of achieving better performance.


\subsection{Implementation of Exploration Reannealing}

As in Q-learning, a simple $\varepsilon$-greedy policy is applied while choosing actions to interact with the environment during training. With large exploration rate $\varepsilon$, the agent fails in exploitation and refining its policy, while with small $\varepsilon$, the agent would have a problem in exploration. For example, if we simply pick $\varepsilon=0.01$, the agent soon learns to hover above the ground forever but hesitates to land. Annealing strategy for exploration rate is considered and tried, in which $\varepsilon$ gradually decreases from 1 to 0.01 during say, half of the training episodes, and $\varepsilon=0.01$ for the rest of training time. However, this cannot solve the hovering problem. This annealing strategy adds randomness at the early stage of training, but the pretrain step (in order to fill the memory for experience replay) has already provided the agent enough exploration stored in the memory at the beginning. Even if the agent learns to land occasionally, it prefers hovering for most of the time. This is probably because the neural network is dealing with continuous state space, learning through some unknown states with bad decisions would also affects the values of well-learned states. As a result, the agent again learns to hover forever. 

In order to escape from such hovering local optima, we carefully engineered a reannealing strategy for the exploration rate. The idea is to encourage the agent to explore while it is hovering. We define a variable \texttt{hover} to count the hovering number which starts from 0. Whenever an episode finishes exceeding the time limit (i.e., the maximum 1000 step in an episode), we increase the hovering number by 1. If the next episode finishes within 1000 steps, we halve the hovering number (using integer division). We will reset $\varepsilon$ back to 1 (for fully exploration) and recount if the hovering number reaches 10 (in this case, the agent tends to hover forever). Otherwise, $\varepsilon$ anneals to 0.01 as described above. 

\subsection{Results}
\begin{figure*}[!htb]
    \centering
    \begin{subfigure}[b]{0.485\textwidth}
        \includegraphics[width=\textwidth]{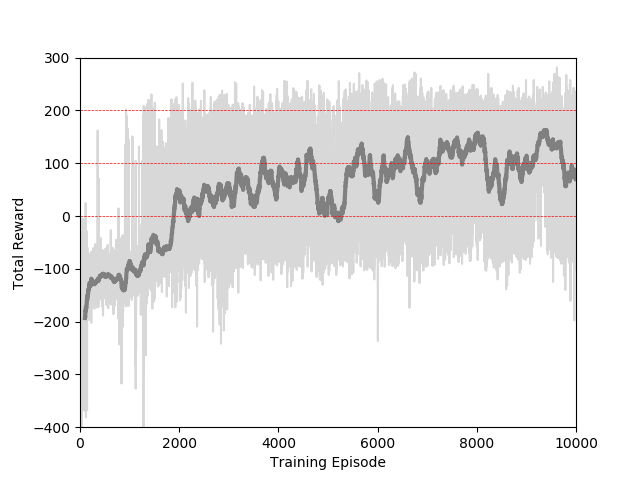}.
        \vspace{-.2in}
        \caption{no reannealing, $\rho_{decay} = 0.99$}
        \label{fig:no99}
    \end{subfigure}
    ~
    \begin{subfigure}[b]{0.485\textwidth}
        \includegraphics[width=\textwidth]{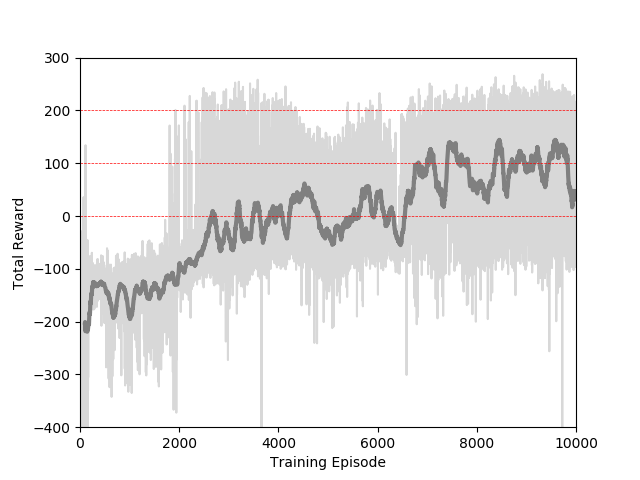}
        \vspace{-.2in}
        \caption{no reannealing, $\rho_{decay} = 0.985$}
        \label{fig:no985}
    \end{subfigure}

    \begin{subfigure}[b]{0.485\textwidth}
        \includegraphics[width=\textwidth]{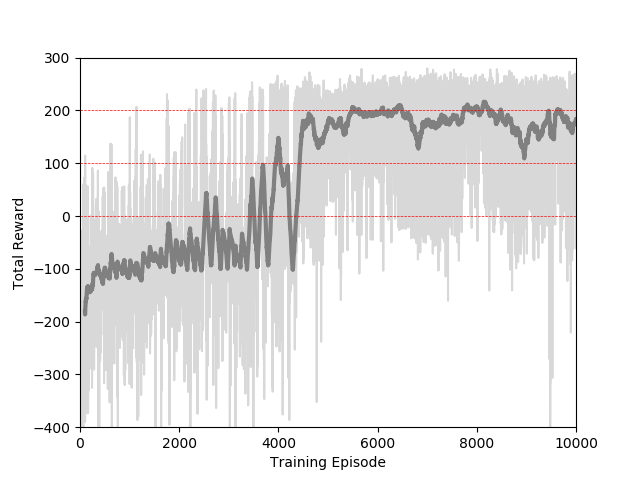}
        \vspace{-.2in}
        \caption{with reannealing, $\rho_{decay} = 0.99$}
        \label{fig:re99}
    \end{subfigure}
    ~
    \begin{subfigure}[b]{0.485\textwidth}
        \includegraphics[width=\textwidth]{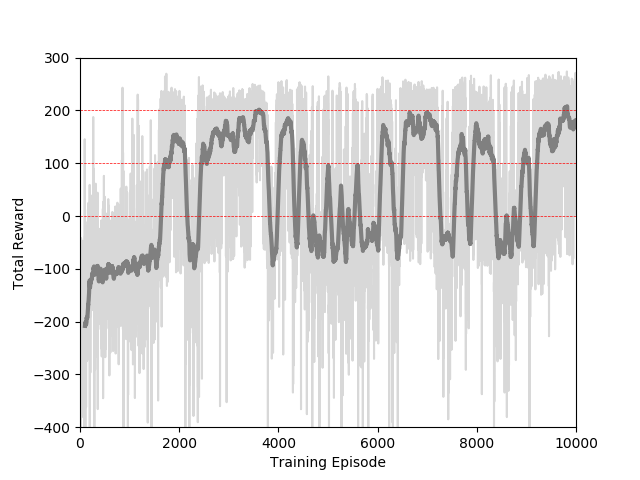}
        \vspace{-.2in}
         \caption{with reannealing, $\rho_{decay} = 0.985$}
        \label{fig:re985}
    \end{subfigure}

    \begin{subfigure}[b]{0.485\textwidth}
        \includegraphics[width=\textwidth]{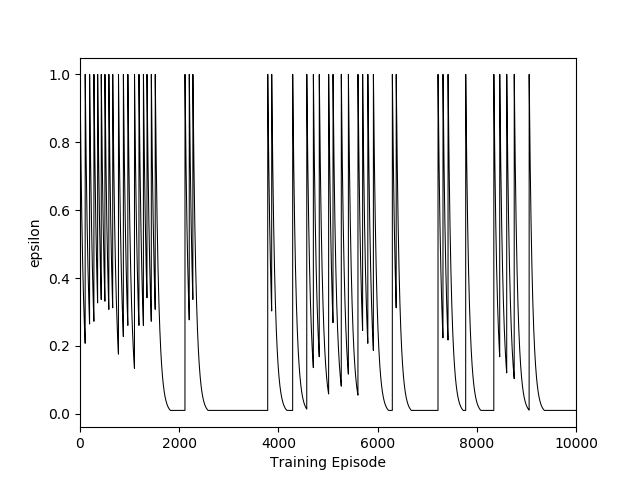}
        \vspace{-.2in}
         \caption{progress of $\varepsilon$, $\rho_{decay} = 0.99$}
        \label{fig:eps99}
    \end{subfigure}
    ~
    \begin{subfigure}[b]{0.485\textwidth}
        \includegraphics[width=\textwidth]{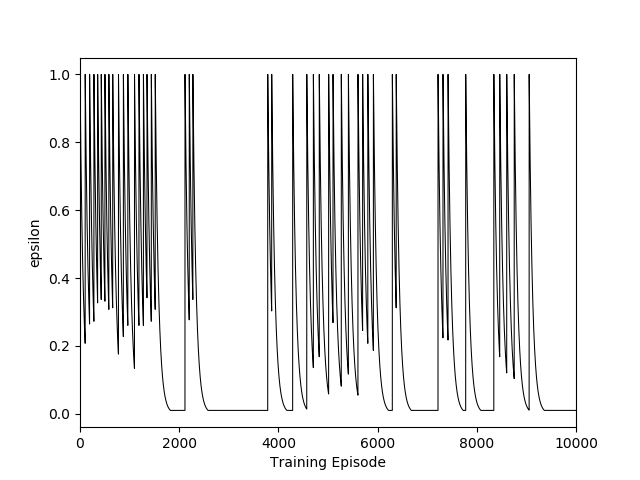}
        \vspace{-.2in}
     \caption{progress of $\rho_{decay} = 0.985$}
        \label{fig:eps985}
    \end{subfigure}

\caption{Performances measured during training. The upper two rows illustrate the total rewards during each episode and moving averages; (a) and (b) correspond to training without reannealing, while (c) and (d) are with exploration reannealing. The bottom row plots the varying $\varepsilon$ values along training with reannealing. In all cases the left column corresponds to exploration decay rate $\rho_{decay}=0.99$, and the right column corresponds to $\rho_{decay}=0.985$}
\label{fig:res}
\end{figure*}

The network was trained over 10,000 episodes. 
Figures \ref{fig:no99} and \ref{fig:no985} 
illustrate the results for the ordinary DQN without applying the exploration reannealing strategy, using two different exploration decay rates. We plot the cumulative rewards for each episode shown with grey, and the smoothed moving averages of the last 100 episodes shown with the blue line. 
Notice that higher rate means slower decay, which results in more exploration at the beginning. In the case that reannealing strategy is not applied, the agent with exploration decay rate $\rho_{decay} = 0.985$ explores less at the beginning than the one with $\rho_{decay} = 0.99$, and performs worse, i.e., its average episodic total rewards are significantly lower, and also the learning process is slower. 
For instance, with $\rho_{decay} = 0.985$, the agent barely learns to avoid crashing (i.e., with episodic rewards above zero) within 3000 episodes, while with $\rho_{decay} = 0.99$, the agent can obtain the same level in about 2000 episodes. Also, to achieve average episodic rewards above 100, it takes less than 4000 episodes with $\rho_{decay} = 0.99$, and more than 7000 episodes with $\rho_{decay} = 0.985$. 
This coincides with our intuition, and emphasizes the importance of sufficient exploration.



As shown in Figures 
\ref{fig:re99} and \ref{fig:re985}, 
applying the reannealing strategy improves our result significantly. 
We could achieve an average value of episodic total reward as high as 200 (in this case, reward 200 means that the agent could land smoothly at the right position on the ground). 
Without reannealing, however, the agent never achieves such level in either cases (see Figures \ref{fig:no99} and \ref{fig:no985}). 
An interesting observation is the steep falls of the moving average along the training while reannealing is applied, clearly these are the moments when $\varepsilon$ is reset to 1. Note that at those times, the falling of episodic total rewards value does not mean the agent is doing worse in general. Q-learning is an off-policy algorithm, which means the learned target policy is not the same as the behavior policy ($\varepsilon$-greedy) it uses while interacting with the environment and accumulating the samples. The 
induced greedy policy has not changed much in such a short period of time from the recent $\varepsilon$ resetting, so the agent can still do as well as before falling if it acts greedily. At the same time, the target policy keeps learning while exploring.
We can see from Figures \ref{fig:re99} and \ref{fig:re985}, 
that for most of the time, it can soon get back to the previous best performance, and often its new peaks are higher, which indicates that it jumps out of the previous local optima. 

We deliberately choose the sliding window size to be not too big, nor do we show the average values over multiple training runs, so that the curves are not over-smoothed, thus allowing us to discern the occurrences of reannealing. 
Over-smoothed curves would give us the illusion that the learning is be slower with reannealing strategy, especially at the early training stages. We claim it is not true, using the same argument that Q-learning is off-policy. We cannot compare the derived policies early on since the exploration rate differ a lot, however, while $\varepsilon$ reaches its minimum value, we can compare the performance of all ``near-greedy'' behavior polices. We see that with reannealing, the agent could reach higher values much faster, thus we claim that reannealing accelerated the training.

We also plot the varying $\varepsilon$ values along training with reannealing strategy in Figures \ref{fig:eps99} and \ref{fig:eps985}, 
from which we can directly observe the moments when reannealing was initiated. There is no need to plot such patterns for the cases  without reannealing, since $\varepsilon$ decays to 0.01 in a few hundred episodes. 
From Figures \ref{fig:eps99} and \ref{fig:eps985} we can see the frequent reannealing early on, since the agent generally can learn to hover very quickly and frequently.
Note that reannealing occurs  more frequently with $\rho_{decay} = 0.985$  than that with $\rho_{decay} = 0.99$. We can surmise then  that our reannealing strategy serves as a remedy for poor hyperparameter tuning, specifically the exploration decay rate, as long as the reannealing criterion (aka the heuristic measure) is appropriately picked. With insufficient exploration at the beginning, the learning would get stuck in poor local optima more often, but reannealing strategy can force the agent to explore later on when it is necessary, and help  find similarly good policy as when training with better hyperparameter. 
Also notice that there is a long flat tail in Figure \ref{fig:eps99} after episode 4500. During this period of training, the agent did not reanneal, and the total reward values stays at that level with smaller variance, compared with no reannealing graphs on Figures \ref{fig:no99} and \ref{fig:no985} .
In fact, we can see that with reannealing, the variance is  smaller when near-greedy policy is applied, i.e., when $\varepsilon$ stays at its minimum for a while. Upon this, we could expect the (greedy) policy learned with reannealing strategy to be superior both in terms of  higher total reward and smaller variance. 


\section{Conclusions} \label{sec:conclude}
In this paper we present a method to organize exploration in RL algorithms. In particular, we focus on its application to model-free value-based approaches, such as DQN. 
Our method is particularly suited to problems which suffer from  poor local optima, and that have sparse rewards as well as long horizons which can trigger the termination criterion earlier. Poor local optima can often be easily distinguished from an outside perspective, yet it may be hard to encode this additional information into reward function or state variables due to complexity of the underlying system. Instead we propose to use a separate, heuristic measure, independent from the agents reward and state, aimed at detecting  the local optima that need to be avoided. 
With such a measure, we can then organize the learning process using a reannealing framework, previously used to solve hard optimization problems.


We highlight some intuitive benefits of applying exploration reannealing, and demonstrate its performance on a standard RL task. In our experiments, reannealing method, indeed helps the agent avoid poor local optima and gather more useful information. The sample efficiency for the reinforcement learning is improved, and the data imbalance problem alleviated. As a result, the training procedure can is accelerated, and the derived policies have superior performance. In addition, we hypothesize that it can serve as a remedy for imperfect hyperparameter tuning.


It is worth noting that the simple framework presented here can be extended to use more sophisticated supervised learning-based heuristic measures for reannealing initiation. If trained properly, such strategies can result in even better performance, due to improved timing of reannealing.

\bibliographystyle{unsrt}
\bibliography{main.bbl}




\end{document}